\documentclass[conference]{IEEEtran}
\IEEEoverridecommandlockouts

\usepackage{amsmath,amssymb,amsfonts}
\usepackage{algorithmic}
\usepackage{graphicx}
\usepackage{hyperref}
\usepackage{textcomp}
\usepackage{cleveref}
\crefname{figure}{Fig.}{Figs.}
\Crefname{figure}{Fig.}{Figs.}
\crefname{table}{Table}{Tables}
\Crefname{table}{Table}{Tables}
\usepackage{xcolor}
\def\BibTeX{{\rm B\kern-.05em{\sc i\kern-.025em b}\kern-.08em
    T\kern-.1667em\lower.7ex\hbox{E}\kern-.125emX}}

\usepackage[utf8]{inputenc}
\usepackage[sorting=none]{biblatex}
\usepackage{booktabs}
\usepackage{multirow}
\usepackage{newunicodechar}
\newunicodechar{̀}{`}

\begin{document}

\title{Align-cDAE: Alzheimer's Disease Progression Modeling with Attention-Aligned Conditional Diffusion Auto-Encoder 
}

\author{\IEEEauthorblockN{Ayantika Das}
\IEEEauthorblockA{\textit{Department of Electrical Engineering,} \\
\textit{Indian Institute of Technology Madras,}\\
Chennai, India \\
ee19d422@smail.iitm.ac.in}
\and
\IEEEauthorblockN{Keerthi Ram}
\IEEEauthorblockA{\textit{Sudha Gopalakrishnan Brain Centre,} \\
\textit{Indian Institute of Technology Madras,}\\
Chennai, India \\
keerthi@htic.iitm.ac.in}
\and
\IEEEauthorblockN{Mohanasankar Sivaprakasam}
\IEEEauthorblockA{\textit{Department of Electrical Engineering,} \\
\textit{Indian Institute of Technology Madras}\\
Chennai, India \\
mohan@ee.iitm.ac.in}
}

\maketitle

\begin{abstract}
Generative AI framework-based modeling and prediction of longitudinal human brain images offer an efficient mechanism to track neurodegenerative progression essential for the assessment of diseases like Alzheimer’s. Among the existing generative approaches, recent diffusion-based models have emerged as an effective alternative to generate disease progression images. Incorporating multi-modal and non-imaging attributes as conditional information into diffusion frameworks has been shown to improve controllability during such generations. However, existing methods do not explicitly ensure that information from non-imaging conditioning modalities is meaningfully aligned with image features to introduce desirable changes in the generated images, such as modulation of progression-specific regions. Further, more precise control over the generation process can be achieved by introducing progression-relevant structure into the internal representations of the model, lacking in the existing approaches.

To address these limitations, we propose a diffusion auto-encoder-based framework for disease progression modeling that explicitly enforces alignment between different modalities. The alignment is enforced by introducing an explicit objective function that enables the model to focus on the regions exhibiting progression-related changes. Further, we devise a mechanism to better structure the latent representational space of the diffusion auto-encoding framework. Specifically, we assign separate latent subspaces for integrating progression-related conditions and retaining subject-specific identity information, allowing better controlled image generation. We have experimentally validated the performance of our model by evaluating on Alzheimer’s disease progression generation through various image similarity metrics and region-wise volumetric assessments. These results demonstrate that enforcing alignment and better structuring of the latent representational space of diffusion auto-encoding framework leads to more anatomically precise modeling of Alzheimer’s disease progression.

\end{abstract}


\section{Introduction}
\label{sec:intro}

The assessment of neurodegenerative diseases such as Alzheimer’s relies on longitudinal imaging to track progressive structural brain changes over time. Longitudinal datasets contain information on disease trajectories, which can be modeled using generative AI techniques, enabling applications such as predicting future time-point images \cite{young2024data, xia2021learning}. Further, modeling of the generation process by incorporating conditional information, like age interval, disease state, and other non-image attributes, enables the disease trajectory to be captured effectively.

\begin{figure}[!b]
  \centering
  \includegraphics[width=0.9
  \linewidth]{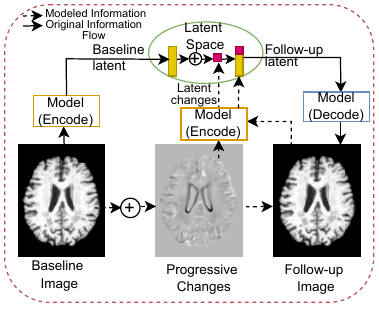}
\caption{Flow diagram indicating the desirable information to be modeled in the latent representational space of our approach.}
\label{fig:abstraction}
\end{figure}

\begin{figure*}[t]
  \centering
  \includegraphics[width=1
  \linewidth]{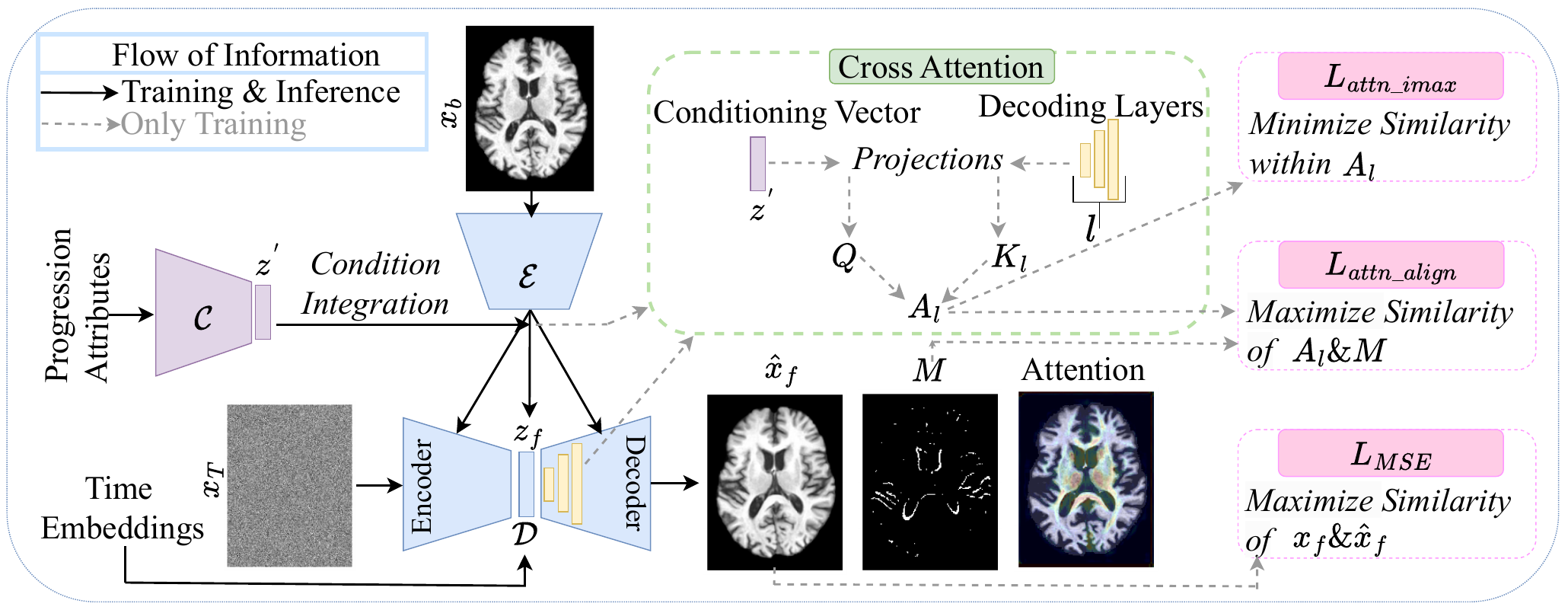}
\caption{Left to right: The condition encoder ($\mathcal{C}$) integrates progression information with the latent representation of baseline image ($x_b$) produced by the encoding component ($\mathcal{E}$), guiding the denoising decoder ($\mathcal{D}$) to generate follow-up image ($\hat{x}_f$). Cross-attention ($A_l$) is computed between the conditioning vector and decoder layers, and the objective functions enforce alignment of $A_l$ with the progression-specific mask ($M$).}
\label{fig:architecture}
\end{figure*}

While Generative Adversarial Networks (GANs) have been utilized in literature for such modeling \cite{ravi2022degenerative, jung2021conditional, xia2021learning, jung2023conditional, wang2023spatial}, recent diffusion-based generative models have emerged as an effective alternative, with a multi-step formulation that iteratively captures image semantics and achieves better fidelity of generated images \cite{ho2020denoising, song2020denoising}. 
Conditional latent diffusion-based models, such as BrLP~\cite{puglisi2024enhancing} and MRExtrap~\cite{kapoor2024mrextrap}, have been explored for modeling disease progression. These approaches integrate multi-modal conditioning in the latent space through feature injection models like ControlNet \cite{zhang2023controlnet}. Although effective in modeling progression patterns, the latent diffusion models are often limited by the reconstruction capability of the auto-encoder that generates the latents \cite{hoogeboom2025simpler, elata2025novel}. To improve spatial control during generation, image-space/ pixel-space diffusion techniques have been employed for progression modeling \cite{yoon2023sadm, litrico2024tadm}. However, these existing approaches offer limited means to explicitly \textit{\textbf{structure}} the \textit{\textbf{internal representations}} of the model and better control the conditional image generation process.

Diffusion-based generative models use a restrictive mechanism of incrementally approaching a possibly complex target distribution from Gaussian noise. The integration of multi-modal and non-image conditioning in diffusion models is usually done by feature space injection (e.g., concatenation/cross-attention or global classifier-free guidance) while optimizing the standard denoising objective \cite{guo2025maisi}. This formulation does not explicitly enforce that the condition induces localized and semantically consistent changes in the generated images \cite{shen2024rethinking, liu2024towards}. How do we \textit{\textbf{better align}} progression \textit{\textbf{conditions with image}} features such that (i) the attention of the model is specific to progression-related regions and (ii) the model is capable of generating precise anatomical changes?

To address this, we propose an image-space diffusion auto-encoder framework that enforces alignment between conditioning (progression-related attributes) and image feature maps and enables the attention of the model to focus on regions exhibiting progression-related changes. 
Furthermore, we devise a mechanism for better structuring the latent representational space of diffusion auto-encoder. Although the diffusion auto-encoding formulation yields a compact and semantically meaningful latent representation space \cite{preechakul2022diffusion, hudson2024soda}, in order to precisely control the progression-related changes, the latent space requires better structuring. As demonstrated in \cref{fig:abstraction}, we have designed an approach that enables the separation of progression-related details from subject-identity information in the latent representations, allowing condition integration in a specific latent subspace and leading to better controllability during image generation. Our contributions are as follows:

\begin{enumerate}
    \item We introduce Align-cDAE, a conditional image-space diffusion framework for generating longitudinal progression images via an explicit \textbf{\textit{alignment}} objective that enforces the model to focus on progression-related anatomical changes.
    
    \item We devise a mechanism to \textit{\textbf{structure}} the latent representational space of a \textit{\textbf{diffusion auto-encoder}} by incorporating progression-related attributes into a dedicated latent subspace, while preserving subject-identity-related information in the remaining subspace.

    \item We demonstrate the efficacy of our approach on Alzheimer’s disease progression MRI datasets, using image similarity metrics and region-wise volumetric evaluations along with ablation studies, showing that alignment-specific enforcement effectively directs the attention of the model toward progression-specific anatomical changes.

\end{enumerate}

\section{Methodology}
We introduce Align-cDAE, a conditional diffusion auto-encoder-based approach for longitudinal disease progression modeling, explicitly enforcing alignment of progression conditions with the feature representations of the model to better capture progression-related changes.
The alignment is enforced through (i) cross-attention estimation between the conditioning attributes and layer-wise feature representations, and further (ii) constraining the cross-attention to emphasize on structural brain changes associated with disease progression. The conditional diffusion auto-encoder and the alignment mechanisms are detailed below in the Subsections \ref{subsec:cDAE} and \ref{subsec:alignment}.
The architectural flow of our method is given in \cref{fig:architecture}.

\subsection{Conditional Diffusion Auto-encoder (cDAE)}
\label{subsec:cDAE}
Our progression-modeling framework is built on a diffusion auto-encoding (DAE) architecture \cite{preechakul2022diffusion} composed of an encoding component ($\mathcal{E}$) and a denoising decoder component ($\mathcal{D}$). The model learns the underlying image distribution by progressively transforming pure Gaussian noise ($x_T  \in \mathbb{R}^{H \times W}$) toward the target image ($x_0  \in \mathbb{R}^{H \times W}$) through a time-dependent denoising process, guided by the encoded representation from $\mathcal{E}$. 
This formulation encourages $\mathcal{E}$ to learn a latent space that captures semantically meaningful image features, while $\mathcal{D}$ reconstructs fine-scale and high-frequency anatomical details \cite{hudson2024soda}. The resulting latent representational space is compact and encapsulates high-level image semantics, enabling organization and separation of progression-related information from subject-specific details. We utilize this latent space to allocate a specific latent subspace for condition incorporation and the remaining latent dimensions for identity-related details.


The goal of the \textit{\textbf{denoising component}} is to iteratively model the reverse denoising process ($p_{\phi}(x_{t-1}|x_{t})$), estimating $x_{t-1}$ by predicting $\hat{x}_{0} = \mathcal{D}( {x}_t, t)$, given $x_t$ at each diffusion step $t$. The noisy image $x_t$ is estimated through the forward diffusion process approximated as, $x_t = \sqrt{\alpha_t } x_0 + \sqrt{1 - \alpha_t} \epsilon$, where $\epsilon \sim \mathcal{N}(0, I)$ and $\alpha_t$ is the noise coefficient. The reverse process to be modeled ($p_{\phi}(x_{t-1}|x_{t})$) is approximated,

\begingroup
\small
\begin{equation}
    \label{eq:q_xt-1_xt_x0}
    q(x_{t-1} | x_t, \hat{x}_0) = \mathcal{N} \left(\sqrt{\alpha_{t-1}} \hat{x}_0  + \sqrt{1 - \alpha_{t-1}} (\frac{x_t - \sqrt{\alpha_t} \hat{x}_0 }  {\sqrt{1 - \alpha_t}}),0 \right)
\end{equation} 
\endgroup

The decoding component adopts a U-Net architecture following the standard DAE implementation \cite{preechakul2022diffusion}.
The \textit{\textbf{encoding component}} ($\mathcal{E}$) maps baseline images ($x_{b}$) into latent representations producing embeddings ($z_b = \mathcal{E}(x_b), z_b \in \mathbb{R}^{d}$), which guide the denoising decoding process ($\hat{x}_{b} = \mathcal{D}( {x}_t, t, \mathcal{E}(x_b))$), formulating an auto-encoding mechanism.
%
Architecturally, the encoding component ($\mathcal{E}$) is similar to the encoder of $\mathcal{D}$, excluding the time-dependency.

\textit{\textbf{Conditioning}}: To generate disease-progressed follow-up images ($x_f$) from a baseline image ($x_b$), the latent representation ($z_b$) extracted by $\mathcal{E}(x_b)$ is conditioned on progression attributes (age ($v_a$) and disease state ($v_d$)).
These attributes are encoded into a conditioning vector ($z^{'}$) using a condition encoder ($\mathcal{C}$), defined as $z^{'} = \mathcal{C}(v_a,v_d)$. Architecturally, $\mathcal{C}$ comprises fully connected layers that map the one-hot encoded progression attributes (more details are in \href{https://github.com/ayantikadas/Align_cDAE/blob/main/Align_cDAE_Supplementary.pdf}{Supplementary}) to $z^{'}$.
The conditioning vector ($z^{'}$) lies in a lower-dimensional latent subspace $z^{'} \in \mathbb{R}^{d^{'}}, d^{'} < d$, as compared to the baseline latent $z_b \in \mathbb{R}^{d}$.
This is further utilized to generate the follow-up latent using,
\begin{equation}
    z_f = z_b + [z^{'}; \mathbf{0}], \mathbf{0} \in \mathbb{R}^{d - d^{'}}
\end{equation}
This design assigns a certain subspace of latent representation ($z_f[0:d^{'}]$) to progression-specific variations, while preserving subject identity in the remaining dimensions $z_f[d^{'}:d]$. The subject-specific information will be obtained from baseline since the latent components are retained ($z_f[d^{'}:d]=z_b[d^{'}:d]$). The follow-up image is then generated using, $\hat{x}_{f} = \mathcal{D}(x_t, t, z_f)$.
%
%

\textit{\textbf{Training and Inference}}: During training, the conditional DAE (cDAE) model is optimized over $T$ diffusion steps while enforcing alignment of the model's feature representations with the progression conditions (Align-cDAE), thereby focusing on anatomical regions affected by disease progression. 
This alignment strategy is detailed in Subsection~\ref{subsec:alignment}. During inference, Align-cDAE predicts follow-up images $\hat{x}_f$ through an iterative denoising process across $T_s$ diffusion steps, guided by the latent representation of the baseline image $x_b$ and conditioned by the progression attributes encoded into the conditioning vector.


\subsection{Alignment Mechanism}
\label{subsec:alignment}


The alignment is achieved by (i) computation of cross-attention between the conditioning vector $z^{'}$ and the layer-wise feature representations ($l \in [1,2,..L]$) in the decoder of $\mathcal{D}$ \cite{vilouras2025anatomy}, and (ii) constraining this attention to emphasize regions exhibiting progression-specific changes. The cross-attention mechanism and the associated objective functions for constraining the attention are detailed in Subsections~\ref{subsubsec:cross-attn} and~\ref{subsubsec:obj-func}.


\subsubsection{Cross Attention}
\label{subsubsec:cross-attn}

Cross-attention is computed between the conditioning vector $z^{'} \in \mathbb{R}^{1 \times d^{'}}$ and the first three decoder layers $K_{l} \in \mathbb{R}^{h \times w \times d_{k}}$ of $\mathcal{D}$, where $d_{k}$, $h$, and $w$ denote the feature depth and spatial dimensions. The vector $z^{'}$ is projected to match the dimensionality of the decoder features, producing $Q \in \mathbb{R}^{d^{'} \times d_{k}}$, using fully connected layers. Cross-attention for each layer is then computed as $A_{l} = (QK_{l}^{T}) / \sqrt{d_{k}}$, yielding $A_{l} \in \mathbb{R}^{d^{'} \times s}$ with $s = h \times w$. The resulting attention maps are subsequently normalized and passed through a softmax function.

\subsubsection{Objective functions}
\label{subsubsec:obj-func}
To optimize Align-cDAE, the layer-wise cross-attention maps $A_{l}$ are aligned to a progression-specific mask that highlights differences between the follow-up $x_f$ and the baseline images $x_b$. While enforcing $A_{l}$ to match this progression mask, diversity within the attention maps needs to be preserved to maximize information content within feature representations. In addition to these, the original DAE reconstruction objective is retained through an MSE loss, $\mathcal{L}_{MSE} = | x_f - \hat{x}_{f} |_{2}^{2}$. The attention alignment and information maximization objectives are detailed in the following subsections.

\textit{\textbf{Attention Alignment}}: The attention alignment is enforced by matching the layer-wise cross-attention maps $A_{l}$ to a progression-specific mask $M$ that highlights changes associated with disease progression.
The mask $M$ is constructed by computing the residual between the baseline image $x_b$ and the follow-up image $x_f$. For similarity maximization, the alignment objective maximizes cosine similarity between $A_{l}$ and $M$, which requires both to be spatially compatible. Thus, $A_{l}$ is averaged across the $d^{'}$ channels and reshaped from $s$ to $h \times w$, yielding $A_{l}^{'} \in \mathbb{R}^{h \times w}$, while $M$ is down-sampled to $M^{'}$ $\in \mathbb{R}^{h \times w}$. The resulting alignment loss is defined as

\begin{equation}
    \label{eq:attn_align}
    \mathcal{L}_{attn\_align} = \frac{1}{L}\left(\sum_{l = 1}^{L}\left( 1 - \cos(A_l^{'}, M^{'})\right)\right)
\end{equation}

where $L$ denotes the number of decoder layers considered for attention alignment. 


\textit{\textbf{Attention Information Maximization}}: The information content within the cross-attention maps $A_l$ is enhanced by encouraging diversity across their channel dimensions. This is achieved by reducing the squared cosine similarity between different channels of $A_l$, leading to the following objective formulation,

\begin{equation}
    \label{eq:attn_imax}
    \mathcal{L}_{attn\_imax} =\frac{1}{L}
    \left(\sum_{l = 1}^{L}\sum_{i=1}^{d^{'}}\sum_{\substack{j=1 \\ j \neq i}}^{s}
    \cos ^ {2}( A_{l}^{\,i}, A_{l}^{\,j})\right)
\end{equation}

%
Hence, the overall objective function is as follows, $\mathcal{L} = \lambda_{1} \mathcal{L}_{attn\_imax}$ $+$ $\lambda_{2} \mathcal{L}_{attn\_align} $ $+$ $\lambda_{3} \mathcal{L}_{MSE}$, where $\lambda_{1}, \lambda_{2}$ and $\lambda_{3}$ are weights to the losses.


\section{Experimental Setup }

\subsection{Dataset and Evaluation Metrics} 
We conduct experiments using longitudinal T1-weighted brain MRI scans from Alzheimer’s Disease Neuroimaging Initiative (ADNI) \cite{jack2008alzheimer}, including subjects diagnosed as cognitively normal (CN), mildly cognitively impaired (MCI), or with Alzheimer’s disease (AD). Images span ages 63–87 and include all genders. For each subject, baseline-follow-up pairs were constructed, with average interval of 2.93 $\pm$ 1.35 years.

\textit{Image Similarity} Metrics: We assess the similarity between generated follow-up ($\hat{x}_{f}$) and ground-truth follow-up ($x_{f}$) images using: Peak Signal-to-Noise Ratio (PSNR), Structural Similarity Index (SSIM), and Mean Squared Error (MSE).
Further, we evaluate the diversity of the generated images using the Fréchet Inception Distance (FID), which quantifies distribution-level similarity \cite{heusel2017gans}.
\textit{Region-wise Volumetric Metrics}: To evaluate progression modeling in 3D, we compute region-wise Mean Absolute Error (MAE) between $(V_{\hat{X}_f}^{r} - V_{{X}_b}^{r})/(V_{X_b}^{r})$ and $(V_{X_f}^{r} - V_{{X}_b}^{r})/(V_{X_b}^{r})$, where $V_{X}^{r}$ denotes the voxel count of region $r$ ($r \in $ Hippocampus/ Amygdala/ Lateral Ventricular) extracted from each 3D volume ($X$) via segmentation using SynthSeg \cite{billot2023synthseg}. 

\subsection{Implementation Details}
\textit{Dataset Details}: From the ADNI cohort, we curated a \textit{Training set} of 486 subjects (179 CN, 160 MCI, 147 AD) and a \textit{Test set} of 466 subjects (159 CN, 156 MCI, 151 AD). The pre-processed steps followed were skull stripping \cite{isensee2019automated}, affine registration to MNI space, and intensity normalization~\cite{shinohara2014statistical}. 
\textit{Model Details}: The models were implemented in PyTorch version 2.0.1 on an 80 GB NVIDIA A100 GPU and CUDA Version: 12.1. The parameter specifications are detailed in \href{https://github.com/ayantikadas/Align_cDAE/blob/main/Align_cDAE_Supplementary.pdf}{Supplementary}. The code-base is made available at \href{https://github.com/ayantikadas/Align_cDAE}{https://github.com/ayantikadas/Align\_cDAE}.

\textit{Baseline Methods}: We have compared with generative models used for progression modeling, including GAN-based approaches (IPGAN~\cite{xia2021learning}, SITGAN~\cite{wang2023spatial}), VAE-based method (DE-CVAE~\cite{he2024individualized}), and a latent diffusion-based approach with ControlNet conditioning (BrLP~\cite{puglisi2024enhancing}). \textit{Ablated Model:} An ablated version of Align-cDAE trained only with the MSE loss, without alignment, was included as cDAE.

\begin{table}[!t]
  \caption{Quantitative evaluations of Align-cDAE with the baseline methods in terms of image-level metrics (PSNR, SSIM, MSE). Statistical significance ($p<0.01$) is marked with ($^*$).}
  \centering
  \resizebox{1\linewidth}{!}{%
  \begin{tabular}{@{}clll@{}}
    \toprule
    \textbf{Methods} &
      \multicolumn{1}{c}{\textbf{PSNR (dB) ($\uparrow$)}} &
      \multicolumn{1}{c}{\textbf{SSIM ($\uparrow$)}} &
      \multicolumn{1}{c}{\textbf{MSE ($\downarrow$)}} \\ 
    \midrule
     &
      \multicolumn{1}{c}{\textbf{\begin{tabular}[c]{@{}c@{}}CN/\\ MCI \& AD\end{tabular}}} &
      \multicolumn{1}{c}{\textbf{\begin{tabular}[c]{@{}c@{}}CN/\\ MCI \& AD\end{tabular}}} &
      \multicolumn{1}{c}{\textbf{\begin{tabular}[c]{@{}c@{}}CN/\\ MCI \& AD\end{tabular}}} \\ 
    \midrule

    Naive Baseline &
      \begin{tabular}[c]{@{}l@{}}27.25 $\pm$ 2.12/\\ 26.75 $\pm$ 2.07\end{tabular} &
      \begin{tabular}[c]{@{}l@{}}0.93 $\pm$ 0.021/\\ 0.92 $\pm$ 0.021\end{tabular} &
      \begin{tabular}[c]{@{}l@{}}0.0021 $\pm$ 0.001/\\ 0.0024 $\pm$ 0.001\end{tabular} \\ 
    \midrule

    \begin{tabular}[c]{@{}c@{}}IPGAN\cite{xia2021learning}\end{tabular} &
      \begin{tabular}[c]{@{}l@{}}25.86 $\pm$ 2.12/\\ 25.31 $\pm$ 2.13\end{tabular} &
      \begin{tabular}[c]{@{}l@{}}0.92 $\pm$ 0.032/\\ 0.91 $\pm$ 0.034\end{tabular} &
      \begin{tabular}[c]{@{}l@{}}0.0030 $\pm$ 0.001/\\ 0.0034 $\pm$ 0.002\end{tabular} \\ 
    \midrule

    \begin{tabular}[c]{@{}c@{}}BrLP\cite{puglisi2024enhancing}\end{tabular} &
      \begin{tabular}[c]{@{}l@{}}26.71 $\pm$ 1.02/\\ 26.20 $\pm$ 1.14\end{tabular} &
      \begin{tabular}[c]{@{}l@{}}0.79 $\pm$ 0.022/\\ 0.79 $\pm$ 0.025\end{tabular} &
      \begin{tabular}[c]{@{}l@{}}0.0029 $\pm$ 0.001/\\ 0.0030 $\pm$ 0.001\end{tabular} \\ 
    \midrule

    \begin{tabular}[c]{@{}c@{}}DE-CVAE\cite{he2024individualized}\end{tabular} &
      \begin{tabular}[c]{@{}l@{}}27.32 $\pm$ 2.98/\\ 26.99 $\pm$ 2.83\end{tabular} &
      \begin{tabular}[c]{@{}l@{}}0.65 $\pm$ 0.090/\\ 0.63 $\pm$ 0.082\end{tabular} &
      \begin{tabular}[c]{@{}l@{}}0.0023 $\pm$ 0.001/\\ 0.0024 $\pm$ 0.001\end{tabular} \\ 
    \midrule
    
    \begin{tabular}[c]{@{}c@{}}SITGAN\cite{wang2023spatial}\end{tabular} &
      \begin{tabular}[c]{@{}l@{}}28.73 $\pm$ 3.25/\\ 28.09 $\pm$ 3.23\end{tabular} &
      \begin{tabular}[c]{@{}l@{}}{0.94 $\pm$ 0.033}/\\ 0.93 $\pm$ 0.034\end{tabular} &
      \begin{tabular}[c]{@{}l@{}}0.0019 $\pm$ 0.001/\\ 0.0022 $\pm$ 0.001\end{tabular} \\ 
    \midrule

    \begin{tabular}[c]{@{}c@{}}cDAE\\ (Ablated Model)\end{tabular} &
      \begin{tabular}[c]{@{}l@{}}28.83 $\pm$ 4.23/\\ 28.27 $\pm$  4.81\end{tabular} &
      \begin{tabular}[c]{@{}l@{}}0.94 $\pm$ 0.038/\\ 0.93 $\pm$  0.050\end{tabular} &
      \begin{tabular}[c]{@{}l@{}}0.0019 $\pm$ 0.001/\\ 0.0021 $\pm$  0.002\end{tabular} \\ 
    \midrule

    \begin{tabular}[c]{@{}c@{}}Align-cDAE\end{tabular} &
      \textbf{\begin{tabular}[c]{@{}l@{}}29.82$^*$ $\pm$ 3.50/\\ 29.52$^*$ $\pm$ 3.82 \end{tabular}} &
      \textbf{\begin{tabular}[c]{@{}l@{}}0.95$^*$ $\pm$ 0.031/\\ 0.94$^*$ $\pm$  0.034\end{tabular}} &
      \textbf{\begin{tabular}[c]{@{}l@{}}0.0018$^*$ $\pm$ 0.001/\\ 0.0019$^*$ $\pm$  0.001\end{tabular}} \\
    
    \bottomrule
  \end{tabular}%
  }
 \label{table:quanti}
\end{table}
\begin{figure*}[t]
  \centering
  \includegraphics[width=1
  \linewidth]{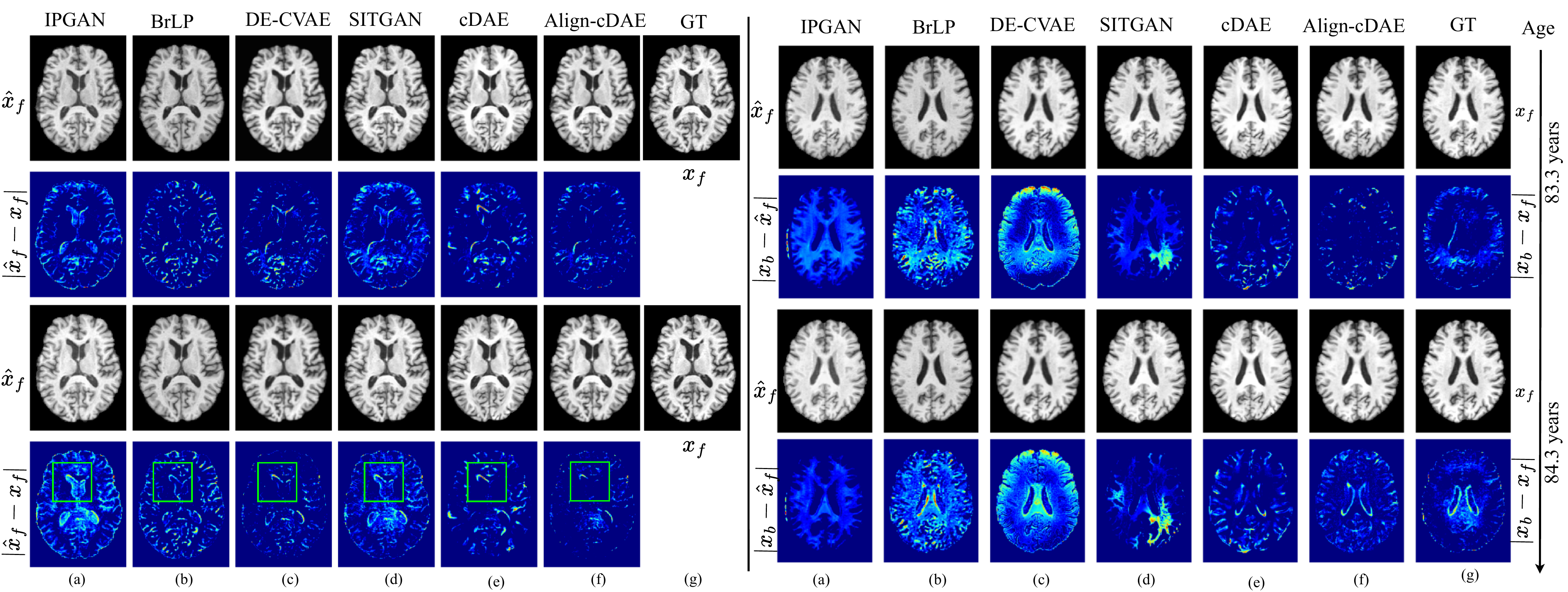}
\caption{Left to right: In sub-figures (A) and (B), the columns (a)-(f) present a method-wise comparison of predicted follow-up images ($\hat{x}_f$) along with their error maps computed with respect to the ground-truth follow-up ($|\hat{x}_f - x_f|$) and ground truth baseline ($|x_b - \hat{x}_f|$), respectively. 
Column (g) in sub-figure (A) shows the ground-truth follow-up (${x}_f$), while in sub-figure (B) it also shows the ground-truth follow-up and baseline difference ($|x_b - x_f|$). 
Top to bottom: For both sub-figures, the upper and lower two rows correspond to subjects aged 83.3 and 84.3 years, respectively, from the AD disease category. In sub-figure (A), the green box highlights that Align-cDAE better predicts progression-related changes. 
}
\label{fig:xf_error}
\end{figure*}



\section{Results and Discussion}


\subsection{Quantitative and Qualitative Analysis}
\label{subsec:Quanti}

The quantitative evaluations comparing Align-cDAE with baseline models are reported in \cref{table:quanti}, using image-level metrics (PSNR, SSIM, and MSE) to compare predicted ($\hat{x}_f$) with ground truth ($x_f$) follow-up image. As shown, Align-cDAE achieves relatively better performance due to the explicit incorporation of the attention alignment mechanism.
As compared to the DAE-based methods, SITGAN performs relatively lower, since the DAE approaches explicitly learn progression-related anatomical changes, while SITGAN relies on an age estimation mechanism to implicitly learn these changes.
Among the remaining baselines, the VAE-based (DE-CVAE) and VAE-latent-diffusion (BrLP) approaches show relatively lower performance, since these models tend to produce blurrier brain structural details, leading to more image-level errors despite the incorporation of better constraints for modeling disease progression.
The GAN-based approach (IPGAN) also employs an implicit mechanism to transform baseline to follow-up images, but lacks awareness about progression-specific changes in the image. 
%
%
Overall, Align-cDAE \textit{\textbf{performs better}} across baselines, demonstrating the effectiveness of conditional alignment that helps to \textit{\textbf{focus}} on \textit{\textbf{progression-specific}} regions for controlled image generations. Further quantification demonstrating the effectiveness of information separation in the latent space is discussed in \href{https://github.com/ayantikadas/Align_cDAE/blob/main/Align_cDAE_Supplementary.pdf}{Supplementary}.


\cref{fig:xf_error} (A) presents a qualitative comparison of absolute error between predicted and ground-truth follow-up images ($|\hat{x}_f - x_f|$) for two future time points (83.3 and 84.3 years) of a subject aged 80.8 years at baseline (AD group).
Align-cDAE shows relatively lower error than all baselines. Compared with the baseline, cDAE, Align-cDAE produces smaller errors in regions associated with disease progression, such as the lateral ventricles. SITGAN exhibits higher error than the DAE-based methods, particularly at 84.3 years, indicating that progression-related factors are not effectively captured. The VAE-based methods (DE-CVAE and VAE–latent-diffusion) capture progression-relevant information but generate blurrier outputs, increasing error in high-frequency and subject-specific brain structures. IPGAN produces higher error, affecting both progression-sensitive and general anatomical areas.
Overall, Align-cDAE better captures progression-specific anatomical changes, \textit{\textbf{reducing error}} in both progression-related and \textit{\textbf{subject identity preserving}} regions.

\begin{table}[!t]
  \caption{Volumetric assessment of Align-cDAE with baseline methods in terms of MAE in different anatomical regions. Statistical significance ($p<0.01$) is marked with ($^*$).}
  \centering
  \resizebox{1\linewidth}{!}{%
  \begin{tabular}{@{}clll@{}}
    \toprule
    \textbf{Methods} &
      \multicolumn{3}{c}{\textbf{MAE ($\downarrow$)}} \\
    \cmidrule(lr){2-4}
    &
      \multicolumn{1}{c}{\textbf{Hippocampus }} &
      \multicolumn{1}{c}{\textbf{Amygdala}} &
      \multicolumn{1}{c}{\textbf{Lateral Ventricles}} \\ 
    \midrule

    IPGAN\cite{xia2021learning} &
      0.3348 $\pm$ 0.0338 &
      0.3283 $\pm$ 0.0260 &
      0.5445 $\pm$ 0.5240 \\ 
    \midrule

    BrLP\cite{puglisi2024enhancing} &
      0.1960 $\pm$ 0.0552 &
      0.1731 $\pm$ 0.0529 &
      0.3702 $\pm$ 0.1012 \\ 
    \midrule

    DE\mbox{-}CVAE\cite{he2024individualized} &
      0.1871 $\pm$ 0.0727 &
      0.1183 $\pm$ 0.1117 &
      0.1747 $\pm$ 0.1470 \\ 
    \midrule

    SITGAN\cite{wang2023spatial} &
      0.1161 $\pm$ 0.0288 &
      0.0211 $\pm$ 0.0201 &
      0.1436 $\pm$ 0.1053 \\ 
    \midrule

    cDAE &
      0.0550 $\pm$ 0.0500 &
      0.0200 $\pm$ 0.0400 &
      0.1250 $\pm$ 0.1200 \\ 
    \midrule

    Align\mbox{-}cDAE &
      \textbf{0.0282$^*$ $\pm$ 0.0203} &
      \textbf{0.0199$^*$ $\pm$ 0.0168} &
      \textbf{0.0705$^*$ $\pm$ 0.0261} \\

    \bottomrule
  \end{tabular}%
  }

\label{tab:vol_analysis}
\end{table}
\textit{Analysis of Progression-related Changes with Age}: \cref{fig:xf_error} (B) compares anatomical changes produced by different methods relative to baseline images (age 80.8 years) for two follow-up time points (83.3 and 84.3 years). The ground-truth difference maps show an age-based progression hierarchy, with more pronounced ventricular expansion at 84.3 years due to accelerated atrophy around the ventricles.
Align-cDAE better reproduces this ventricular growth pattern and preserves the expected progression hierarchy.
%
%
The DAE baseline (cDAE) captures ventricular enlargement but lacks the precision needed to localize progression-specific changes. GAN-based approaches (SITGAN and IPGAN) produce outputs that remain close to the baseline image, not precisely highlighting the expected progression in the ventricular region. VAE and latent-diffusion VAE methods introduce broader, less localized differences both within and outside progression-sensitive regions.
%
Overall, the difference map of Align-cDAE indicates that the design specification introduced to focus on progression-related changes \textit{\textbf{facilitates}} generation of the \textit{\textbf{required anatomical}} changes by preserving subject identity.

\subsection{Region-wise Volumetric Analysis}
\label{vol_analysis}
To evaluate how well progression-related changes are captured in 3D, we compare normalized volumetric changes between predicted $(V_{\hat{X}_{f}})$ and ground-truth follow-up $(V_{X_{f}})$ volumes relative to the baseline $(V_{X_{b}})$. The 2D generated images ($\hat{x}$) are sequentially stacked, $\hat{X}_{:,:,c} = \hat{x}$ to generate the 3D images ($\hat{X} \in \mathbb{R}^{H \times W \times D}, \; c \in \{1,\dots,D\}$). The MAE between $(V_{\hat{X}_{f}}^{r} - V_{X_b}^{r})/(V_{X_b}^{r})$ and $(V_{X_f}^{r} - V_{X_b}^{r})/(V_{X_b}^{r})$ for predicted and ground truth follow-ups respectively are reported in \cref{tab:vol_analysis}, for three anatomical regions ($r$).  
Align-cDAE achieves relatively lower errors across all regions, consistent with the image-level results where DAE-based methods are better than other baselines. GAN-based approaches (SITGAN and IPGAN) rely on implicit age constraints and do not capture region-specific anatomical changes precisely, resulting in higher errors. VAE and latent-diffusion VAE models produce non-localized volumetric differences due to blurrier high-frequency reconstructions, limiting their ability to model precise progression.
%
%
Overall, Align-cDAE shows \textit{\textbf{lower volumetric error}} by concentrating anatomical changes in progression-relevant regions, producing \textit{\textbf{meaningful 3D}} progressions suitable for downstream analyses.

\begin{figure}[!b]
  \centering
  \includegraphics[width=0.75
  \linewidth]{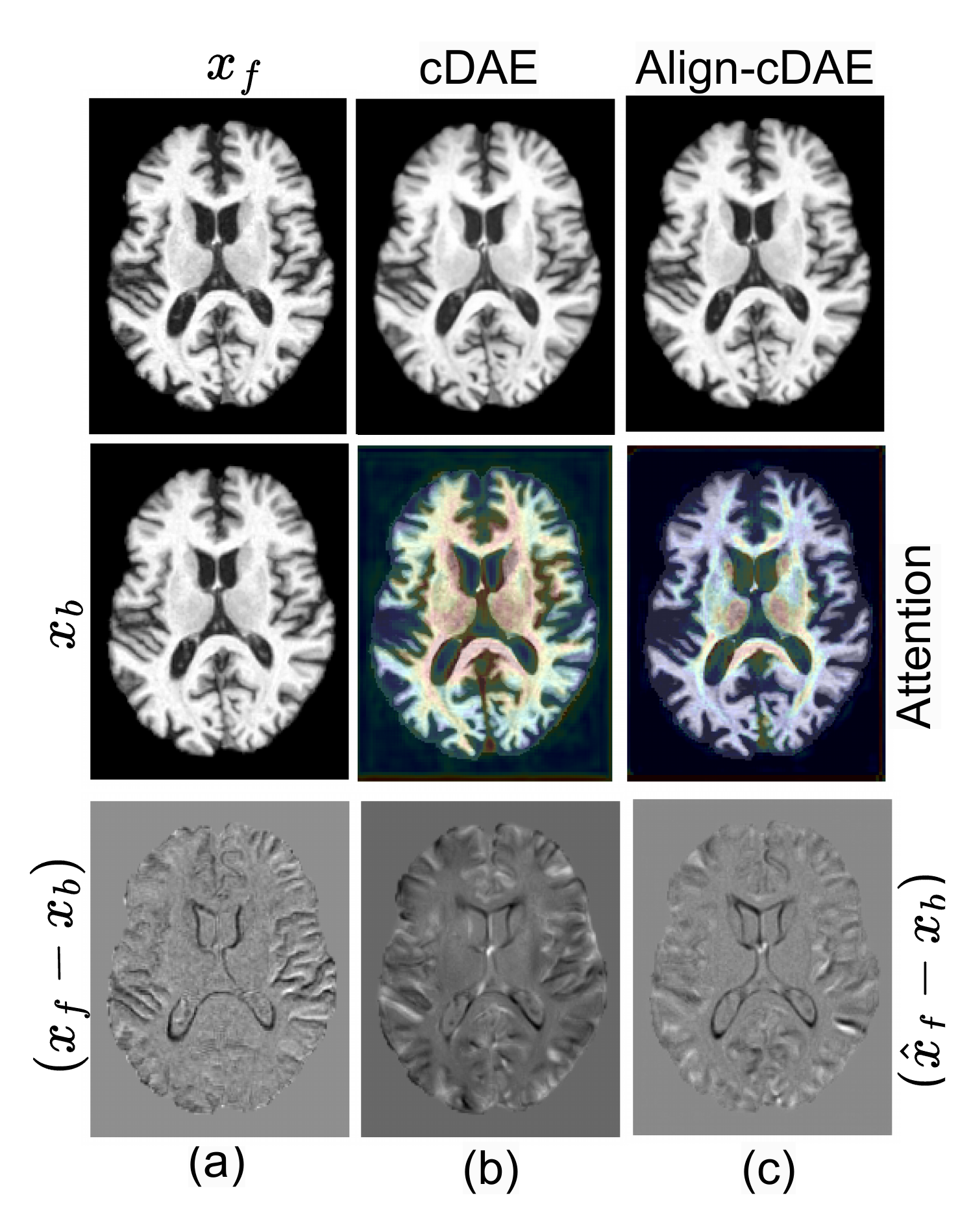}
\caption{Left to right: (a) Ground truth, (b) cDAE, and (c) Align-cDAE. Top to bottom: (i) Ground-truth and predicted follow-up images, (ii) (a) Ground truth baseline and (b)-(c) model attention maps overlaid on images, and (iii) Difference between ground-truth/ predicted follow-up and ground-truth baseline images.}
\label{fig:attention}
\end{figure}

\subsection{Analysis of Attention Alignment}
\label{subsec:attention_align}

In order to assess the impact of attention alignment on our modeling approach, we extract the mid-layer ($l = 10$, not considered for alignment loss) of the decoder of $\mathcal{D}$ and visualize the attention map by averaging across its channel dimension. \cref{fig:attention} compares the attention maps of Align-cDAE and cDAE from layers not considered for loss computation, alongside their corresponding difference maps highlighting generated changes. 
(i) \textit{Attention Maps}: From the figure, it is evident that Align-cDAE focuses attention around progression-relevant regions, particularly the ventricles, whereas cDAE shows a diffuse attention pattern. 
(ii) \textit{Difference Maps}: The difference maps (between the ground-truth baseline and predicted follow-up ($\hat{x}_f - x_b$)) produced by Align-cDAE more closely resemble the actual region-wise changes seen in the ground-truth ($x_f - x_b$). This suggests that Align-cDAE \textit{\textbf{better captures progression}} patterns.

\begin{table}[!ht]
  \caption{Cognitive state classification accuracy under different mixtures of real baseline data (RD) and generated follow-up data (GD).}
  \centering
  \resizebox{0.55\linewidth}{!}{%
  \begin{tabular}{@{}cc@{}}
    \toprule
    \textbf{Data Mixture} & \textbf{Accuracy (\%)} \\
    \midrule
    $100\%$ RD $+ 0\%$ GD   & $81.92$ \\\midrule
    $75\%$ RD $+ 25\%$ GD  & $82.02$ \\\midrule
    $50\%$ RD $+ 50\%$ GD  & $82.15$ \\\midrule
    $25\%$ RD $+ 75\%$ GD  & $82.00$ \\\midrule
    $0\%$ RD $+ 100\%$ GD  & $81.93$ \\
    \bottomrule
  \end{tabular}%
  }
  \label{tab:downstream}
\end{table}

\begin{table}[]
\caption{Ablation study comparing the image similarity and diversity metrics, along with downstream classification performance.}
\centering
\resizebox{0.9\columnwidth}{!}{%
\begin{tabular}{@{}ccccc@{}}
\toprule
\textbf{Methods} &
  \textbf{\begin{tabular}[c]{@{}c@{}}PSNR($\uparrow$)\\(dB)\end{tabular}} &
  \textbf{\begin{tabular}[c]{@{}c@{}}SSIM($\uparrow$)\end{tabular}} &
  \textbf{\begin{tabular}[c]{@{}c@{}}FID($\downarrow$)\end{tabular}} &
  \textbf{\begin{tabular}[c]{@{}c@{}}Accuracy($\uparrow$)\\(\%)\end{tabular}} \\ \midrule
 &
  \multicolumn{3}{c}{\begin{tabular}[c]{@{}c@{}}CN/\\ MCI \& AD\end{tabular}} &
  \begin{tabular}[c]{@{}c@{}}Classify \\ CN vs. AD\end{tabular} \\ \midrule
\begin{tabular}[c]{@{}c@{}}Align-cDAE \\ (w/o L\_attn\_imax)\end{tabular} &
  \begin{tabular}[c]{@{}c@{}}28.58/ \\ 28.54\end{tabular} &
  \begin{tabular}[c]{@{}c@{}}0.945/\\ 0.938\end{tabular} &
  \begin{tabular}[c]{@{}c@{}}7.56/ \\ 7.88\end{tabular} &
  80.99 \\ \midrule
Align-cDAE &
  \begin{tabular}[c]{@{}c@{}}29.82/ \\ 29.52\end{tabular} &
  \begin{tabular}[c]{@{}c@{}}0.950/\\ 0.944\end{tabular} &
  \begin{tabular}[c]{@{}c@{}}6.85/ \\ 6.93\end{tabular} &
  81.93 \\ \bottomrule
\end{tabular}%
}
\label{tab:loss_ablation}
\end{table}

\subsection{Downstream Classification}
\label{downstream}

In order to evaluate the effectiveness of generated follow-up images (GD) relative to real data (RD), a cognitive state (CN vs. AD) classifier (ResNeXt~\cite{xie2017aggregated} with MLP layers) is trained using different RD–GD mixtures from the \textit{Train Set} and evaluated on real data (RD) from the \textit{Test Set}. The classification accuracies are reported in \cref{tab:downstream}.
The experimental design is such that RD is sampled from real baseline (previous time-point) images, and GD is sampled from generated follow-up (follow-up time-point) images, with no overlapping samples.

From \cref{tab:downstream}, it is inferred that using only RD (81.92\%) or only GD (81.93\%) yields nearly similar performance, indicating that the generated follow-up images are informative enough for the downstream task. Further, combining RD and GD while keeping the total training size fixed results in consistent gains, with relatively more accuracy at a $50\%$–$50\%$ mixture (82.15\%). This improvement reflects the benefit of increased diversity from mixing baseline and generated follow-up images. Overall, the generated samples from Align-cDAE can \textit{\textbf{effectively augment}} limited real data scenarios.

\subsection{Ablation Study with Objective Functions}
\label{ablation}
In order to assess the role of information maximization loss ($\mathcal{L}_{attn\_imax}$ in Equation~\ref {eq:attn_imax}), we performed an ablation study without the incorporation of $\mathcal{L}_{attn\_imax}$ and evaluated on the \textit{Test Set}. The comparison results of Align-cDAE with the ablated model (Align-cDAE w/o $\mathcal{L}_{attn\_imax}$) are reported in \cref{tab:loss_ablation}. The tabulated results highlight that the scores of the image similarity metrics (PSNR and SSIM) are relatively lower without the incorporation of the information maximization loss. Specifically, the diversity score reported through FID is affected when the loss is not incorporated, aligning with the formulation of the loss to enhance layer-wise diversity. This consequently affects the downstream classification performance. Overall, the incorporation of the loss enables a \textit{\textbf{balance}} between enforcing attention-based \textit{\textbf{alignment}} and preserving image \textit{\textbf{diversity}}, thereby improving the learning of better anatomical details.

\section{Conclusion}
We have introduced a diffusion auto-encoding framework that enforces alignment between multi-modal conditioning and image features, enabling precise modulation of longitudinal disease progression synthesis. Further, a structuring within the latent representational space of diffusion-auto encoder was incorporated, allowing separation between progression- and subject-identity-related information. We have demonstrated the effectiveness of our approach through evaluating the model against multiple baselines using (i) image similarity, (ii) region-level volumetric assessment, and (iii) downstream analysis on an Alzheimer’s disease progression dataset. Ablation studies further demonstrate that enforcing alignment steers the decoder layers to focus attention on progression-specific anatomical regions.
\textit{Limitations and Future Scope:} The current model incorporates limited progression attributes (age and disease state) for conditioning, which can be extended to richer clinical modalities that could further enhance the modeling of progression patterns. Overall, our findings highlight that it is essential for conditional models to focus on \textit{\textbf{alignment of multi-modal information}} for achieving precise image generation.

\section*{Acknowledgment}

We would like to thank the team at Sudha Gopalakrishnan Brain Centre, IITM, for their consistent support. Data collection and sharing for this project was funded by the Alzheimer's Disease Neuroimaging Initiative (ADNI) (National Institutes of Health Grant U01 AG024904) and DOD ADNI (Department of Defense award number W81XWH-12-2-0012).

{\small
\printbibliography
}
\end{document}